\documentclass[10pt,twocolumn,letterpaper]{article}

\usepackage{cvpr}
\usepackage{times}
\usepackage{graphicx}
\usepackage{amsmath}
\usepackage{amssymb}
\usepackage{mathtools}
\usepackage{amsthm}
\usepackage{subcaption}
\usepackage{paralist}
\usepackage{bm}
\usepackage{amsfonts}
\usepackage{multirow}
\usepackage{cases}
\usepackage{booktabs}
\usepackage{threeparttable}
\usepackage{upgreek}
\usepackage{endnotes}
\usepackage{etoolbox}
\usepackage{algpseudocode}
\usepackage{algorithm}
\usepackage{bbding}
\newtheorem{theorem}{Theorem}
\newcommand{\tabincell}[2]{\begin{tabular}{@{}#1@{}}#2\end{tabular}}

\usepackage[breaklinks=true,bookmarks=false]{hyperref}

\cvprfinalcopy 

\def\cvprPaperID{****} 

\begin{document}

\title{Selective Transfer with Reinforced Transfer Network for Partial Domain Adaptation}
\author{ Zhihong Chen, Chao Chen, Zhaowei Cheng, Boyuan Jiang, Ke Fang, Xinyu Jin\\
 Institue of Information Science and Electronic Engineering, Zhejiang University\\
 \tt\small \{zhihongchen,chench,chengzhaowei,byjiang,ke-fang,jinxy\}@zju.edu.cn}


\maketitle

\begin{abstract}
One crucial aspect of partial domain adaptation (PDA) is how to select the  relevant source samples in the shared classes for knowledge transfer. Previous PDA methods tackle this problem by re-weighting the source samples based on their high-level information (deep features). However, since the domain shift between source and target domains, only using the deep features for sample selection is defective. We argue that it is more reasonable to additionally exploit the pixel-level information for PDA problem, as the appearance difference between outlier source classes and target classes is significantly large. In this paper, we propose a reinforced transfer network (RTNet), which utilizes both high-level and pixel-level information for PDA problem. Our RTNet is composed of a reinforced data selector (RDS) based on reinforcement learning (RL),   which filters out the outlier source samples, and a domain adaptation model which minimizes the domain discrepancy in the shared label space. Specifically, in the RDS, we design a novel reward based on the reconstruct errors of selected source samples on the target generator, which introduces the pixel-level information to guide the learning of RDS. Besides, we develope a state containing high-level information,  which used by the RDS for sample selection. The proposed RDS is a general module, which can be easily integrated into existing DA models to make them fit the PDA situation. Extensive experiments indicate that RTNet can achieve state-of-the-art performance for PDA tasks on several benchmark datasets.
\end{abstract}

\section{Introduction} \label{introduction}
Deep neural networks have achieved impressive performance in a variety of applications. However, when applied to related but different domains, the generalization ability of the learned model may be severely degraded due to the harmful effects of the domain shift \cite{candela2009dataset}. Re-collecting labeled data from the coming new domain is prohibitive because of the huge cost of data annotation. Domain adaptation (DA) techniques solve such a problem by transferring knowledge from a source domain with rich labeled data to a target domain where labels are scarce or unavailable. These DA methods learn domain-invariant features by moment matching \cite{long2017deep,sun2016deep,chen2019deep} or adversarial training \cite{tzeng2017adversarial,ganin2016domain}.
\begin{figure}[!t]
   \centering
     \includegraphics*[width=1.05\linewidth]{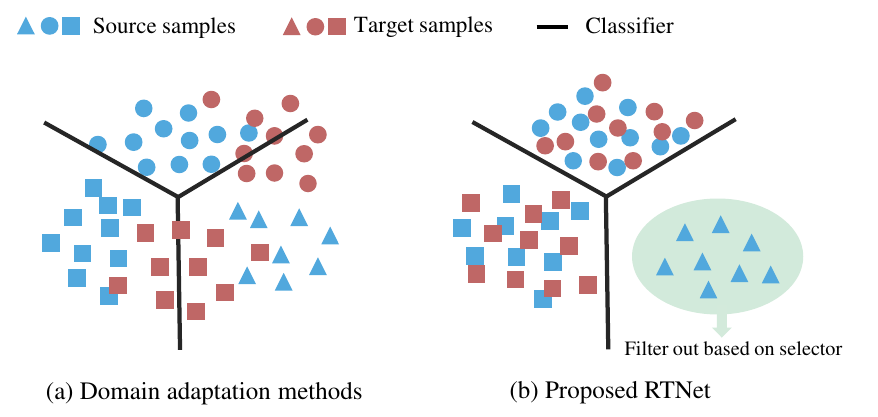}
   \caption{(a) Negative transfer is triggered by mismatch. (b) Negative transfer is mitigated by filtering out outlier classes.}
   \label{Fig1}
\end{figure}

Previous DA methods generally assume that the source and target domains have shared label space, i.e., the category set of the source domain is consistent with that of the target domain. However, in real applications, it is usually formidable to find a relevant source domain with identical label space as the target domain. Thus, a more realistic scenario is partial domain adaptation (PDA) \cite{cao2018partial}, which relaxes the constraint that source and target domains share the same label space and assumes that the unknown target label space is a subset of the source label space. In such a scenario, as shown in Figure \ref{Fig1}a, existing DA methods force an error match between the outlier source class (blue triangle) and the unrelated target class (red square) by aligning the whole source domain with the target domain. As a result, the negative transfer may be triggered due to the mismatch. Negative transfer is a dilemma that the transfer model performs even worse than the non-adaptation (NoA) model \cite{pan2010survey}.

Several approaches have been proposed to solve the PDA problem by re-weighting the source samples, where the weights can be get from the distribution of the predicted target label probabilities \cite{cao2018partial2} or the prediction of the domain discriminator \cite{cao2018partial,zhang2018importance}. These methods select the relevant source  samples only considering the high-level information (deep features), which however ignore the most discriminative features hidden in the pixel-level information, such as appearance, style or background. Since the difference of appearance between the outlier source samples and the target samples is significantly large, taking into account the pixel-level information for outlier sample selection is expected to benefit the adaptation performance \cite{hoffman2018cycada}. Moreover, these PDA modules based on adversarial networks are difficult to integrate into matching-based DA methods lacking discriminators. Therefore, most existing matching-based methods are hard to extend to address the PDA problem.

In this paper,  to address the PDA problem, we present a reinforced transfer network (RTNet), as shown in  Figure \ref{Fig1}b,  which exploits reinforcement learning (RL) to learn a reinforced data selector (RDS) for  filtering outlier source samples. In this respect, the DA model from RTNet can align distributions in the shared label space to avoid negative transfer.  To utilize both pixel-level and high-level information, we design a RDS. The RDS takes action (keep or drop a sample) based on the state of sample. Then, the reconstruction error of the selected source sample on the target generator is used as a reward to guide the learning of RDS via the actor-critic algorithm \cite{konda2000actor}. Note that, the state contains high-level information, and the reward contains pixel-level information. Specifically, the intuition of using reconstruction error to introduce pixel-level information is that the target generator lacks training samples of outlier classes and the outlier source samples  extremely dissimilar to the target classes, so on the generator trained with target samples, the reconstruction error of outlier sample is larger than that of the  related source samples. Hence, the reconstruction error can measure the appearance similarity between each source sample and the target domain well, which is the important information in sample selection and hard to get from high-level information.

The contributions of this work are: (1) a novel PDA framework RTNet is proposed, which joints sample selection and domain discrepancy minimization. (2) we design a reinforced data selector based on reinforcement learning,  which solves the PDA problem by taking into account high-level and pixel-level information to select related samples for positive transfer. As far as we know, this is the first work to address PDA problem with RL technique. (3) most DA methods can be extended to solve PDA problem by integrating the RDS. We use two types of base network to evaluate the effectiveness of integration. (4) The RTNet achieves the best performance on three well-known benchmarks.

\section{Related Work}
\textbf{Partial Domain Adaptation:} Deep DA methods have been widely studied in recent years. These methods extend deep models by embedding adaptation layers for moment matching \cite{tzeng2014deep,long2015learning,sun2016deep,chen2019joint,chen2019homm} or adding domain discriminators for adversarial training \cite{ganin2016domain,tzeng2017adversarial}. However, these methods may be restricted by the assumption that source and target domains share the same label space, which is not held in PDA scenario. Several methods have been proposed to solve the PDA problem. Selective adversarial network (SAN) \cite{cao2018partial} trains a separate domain discriminator for each class with a weight mechanism to suppress the harmful influence of outlier classes. Partial adversarial domain adaptation (PADA) \cite{cao2018partial2} improves SAN by adopting only one domain discriminator and gets the weight of each class based on the target probability distribution output by classifier. Example transfer network (ETN) \cite{cao2019learning} quantifies the weights of source examples based on their similarities to the target domain. Unlike previous PDA methods, only high-level information was used, RTNet combines pixel-level and high-level information to achieve more accurate sample filtering.

\textbf{Reinforcement Learning:} RL can be roughly divided into two categories \cite{arulkumaran2017deep}: value-based methods and policy-based methods. The value-based methods estimate future expected total rewards through a state, such as SARSA \cite{rummery1994line} and deep Q network \cite{mnih2015human}.  Policy-based methods try to directly find the next best action in the current state, such as REINFORCE algorithm \cite{williams1992simple}. To reduce variance, some methods combine value-based and policy-based methods for more stable training, such as the actor-critic algorithm \cite{konda2000actor}. So far, data selection based on RL has been applied in the fields of active learning \cite{fang2017learning}, co-training \cite{wu2018reinforced}, text matching \cite{qu2019learning}, etc. However, there is a lack of reinforced data selection methods to solve the PDA problem.
\begin{figure*}[ht]
   \centering
     \includegraphics*[width=0.83\linewidth]{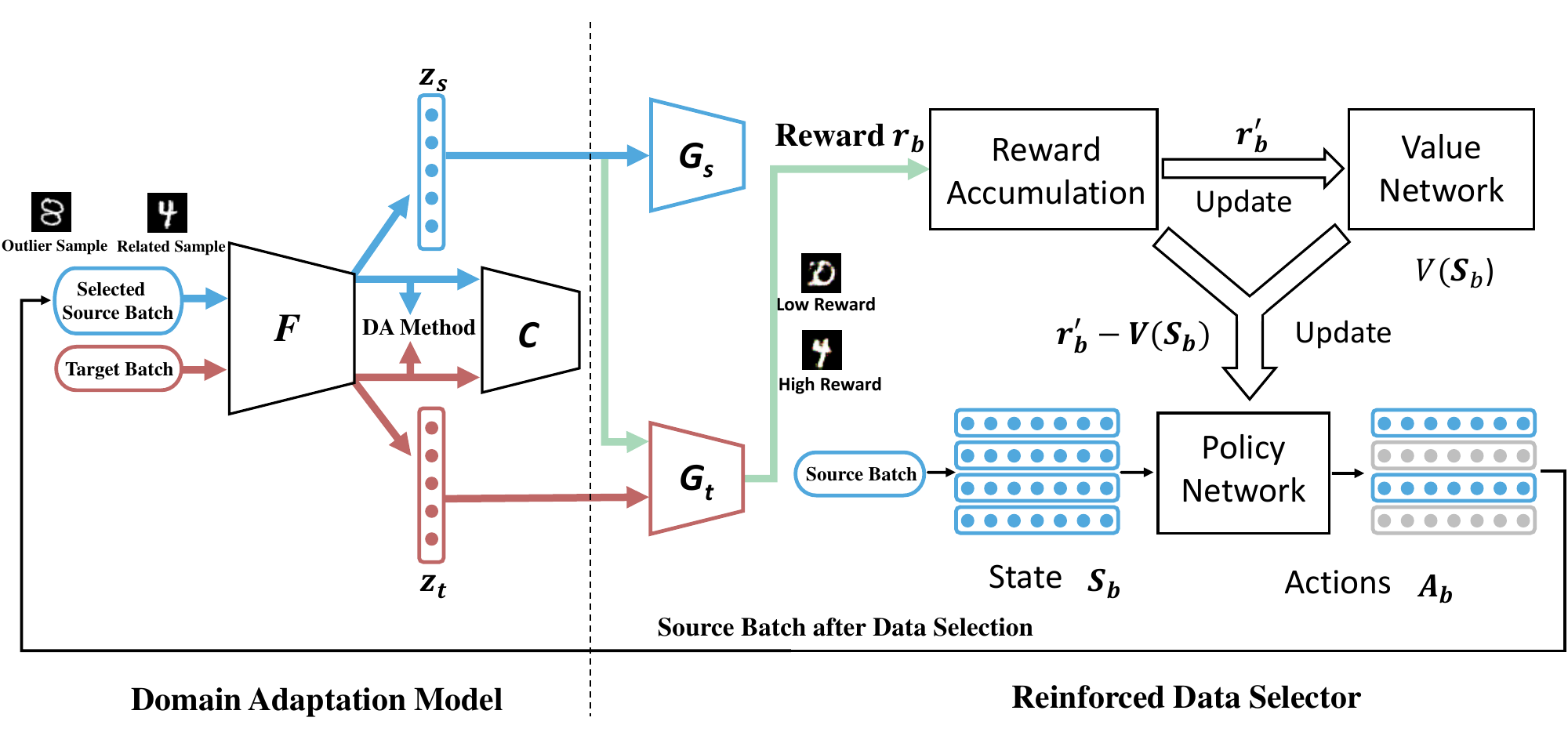}
     \caption{ Overview of RTNet. $F$ is a shared feature extractor, $C$ is a shared classifier, $G_s$ and $G_t$ are source and target generators respectively,  $V$ is a value network and $\pi$ is a policy network. $G_s$ and $G_t$ are combined with $F$ to construct source and target auto-encoders to reconstruct samples, respectively. The green line indicates the flow to get the reward.}
     \label{Fig2}
\end{figure*}
\section{Our Approach} \label{section3}
\textbf{Problem Definition and Notations:}
 In this work, based on PDA settings, we define the labeled source dataset as $\{\bm{X}^s,Y^s\}=\{(\bm{x}_i^s,y_i^s)\}_{i=1}^{n_s}$ from source domain $\mathcal{D}_s$ associated with $|\mathcal{C}_s|$ classes, and define the unlabeled target dataset as $\{\bm{X}^t\}=\{\bm{x}_i^t\}_{i=1}^{n_t}$ from target domain $\mathcal{D}_t$ associated with $|\mathcal{C}_t|$ classes. Note that, the target label space is contained in the source label space, i.e., $\mathcal{C}_t\in\mathcal{C}_s$ and $\mathcal{C}_t$ is unknown. The two domains follow different marginal distributions, $p$ and $q$, respectively, we further have $p_{\mathcal{C}_{t}} \neq q$. $p_{\mathcal{C}_{t}}$ is the distribution of source samples in the target label space. The goal is to improve the performance of model in $\mathcal{D}_t$ with the help of the knowledge in $\mathcal{D}_s$ associated with $\mathcal{C}_t$.
 \subsection{Overiew of RTNet}
As shown in Figure \ref{Fig2}, RTNet consists of two components: a domain adaptation model ($F$ and $C$) and a reinforced data selector ($G_{s,t}$, $V$ and $\pi$). The DA model promotes positive transfer by reducing distribution shift in the shared label space. The RDS based on RL mitigates negative transfer by filtering outlier source classes. Specifically, to filter outlier source samples, the policy network $\pi$ considers high-level information provided by feature extractor $F$ and classifier $C$ for decision making to get selected source samples $\bm{X}^{s'}$. For the backbone of DA model, $C$ takes source transfer features $\bm{Z}^{s'}=F(\bm{X}^{s'})$ as input to produce label predictions $\hat{Y}^s$, and $F$ achieves distribution alignment between $F(\bm{X}^{s'})$ and $F(\bm{X}^{t})$. Meanwhile, the selected source samples' reconstruction errors $\|\bm{X}^{s'}-G_{t}F(\bm{X}^{s'})\|_2^2$ based on $G_t$ are used as a reward to encourage $\pi$ to select samples with small reconstruction errors.  For the stability of training, based on actor-critic algorithm, we use a value network $V$ combined with rewards to optimize $\pi$. Besides, the domain-specific generators $G_s$ and $G_t$ trained with reconstruction errors of reconstructed source images $G_{s}(F(\bm{X}^s{'}))$ and target images $G_{t}(F(\bm{X}^t))$, respectively.

\subsection{Domain Adaptation Model}
Almost all PDA frameworks are based on adversarial network \cite{cao2018partial,cao2018partial2,zhang2018importance,cao2019learning}, which has led to many existing DA algorithms based on moment matching cannot be extended to solve the PDA problem. The proposed RDS is a general module that can be integrated into most DA frameworks. In this work, we use deep CORAL\cite{sun2016deep} as the base DA model to prove that RDS can be embedded into the matching-based DA framework to make it robust to PDA scene. The reason why we chose CORAL is that it is simple and effective. Although there are some limitations in CORAL,  it is beyond our research scope. Besides,  the RDS is universal, so CORAL can be replaced with other better DA methods. In the Appendix, we also provide a solution for embedding RDS into DANN to demonstrate that RDS can also be integrated into the method based on adversarial network. In the following, we will give a brief introduction of CORAL.

We define the last layer of $F$ as adaptation layer and reduce the distribution shift between source and target domains by aligning the covariance matrix of source and target features. Hence, the CORAL objective function is:
\begin{equation}
\mathop{\mathcal{L}_{c}}\limits_{(F)}=\frac{1}{d^2}\|Cov(\bm{Z}_b^s)-Cov(\bm{Z}_b^t)\|_F^2,
\end{equation}
where $\|\cdot\|_F^2$ denotes the squared matrix Frobenius norm, $\bm{Z}_{b}^s \in \mathbb{R}^{n\times d}$ and $\bm{Z}_{b}^t \in \mathbb{R}^{n\times d}$ represent source and target transferable features output by the adaptation layer, respectively, $b$ is the batch ID, $d$ is the dimension of the transferable feature, and n is the batch size. $Cov(\bm{Z}_b^s)$ and $Cov(\bm{Z}_b^t)$ represent the covariance matrices, which can be computed as $Cov(\bm{Z}_b^s)=\mathbf{Z}_b^{s\top}\mathbf{Z}_b^s$, and $Cov(\mathbf{Z}_b^t)=\mathbf{Z}_b^{t\top}\mathbf{Z}_b^t$.

To ensure the shared feature extractor and classifier can be trained with supervision on labeled samples, we define a standard cross-entropy classification loss $\mathcal{L}_{s}$  with respect to labeled source samples. Formally, the full objective function for the domain adaptation model is as follows:
\begin{equation}
\mathcal{L}_{DA}=\mathcal{L}_{s}+\lambda_1 \mathcal{L}_{c},
\end{equation}
where hyperparameter $\lambda_1$ control the impact of the corresponding objective function. However, in the PDA scenario, most DA methods (e.g. CORAL) may trigger negative transfer since these methods force alignment of the global distributions $p$ and $q$, even though $p_{\mathcal{C}_{s} \setminus \mathcal{C}_{t}}$ and  $q$ are non-overlapping and cannot be aligned during transfer. Thus, the motivation of the reinforced data selector is to mitigate negative transfer by filtering out the outlier source classes $\mathcal{C}_{s}\setminus \mathcal{C}_{t}$ before performing the distribution alignment.
\subsection{Reinforced Data Selector}
We consider the source sample selection process of RTNet as Markov decision process, which can be addressed by RL. The RDS is an agent that interacts with the environment created by the DA model. The agent takes action to keep or drop a source sample based on the policy function.  The DA model evaluates the actions taken by the agent and provides a reward to guide the learning of agent.

As shown in  Figure \ref{Fig2}, given a batch of source samples $\bm{X}_b^{s}=\{\bm{x}_i^s\}_{i=1}^n$, we can obtain the corresponding states $\bm{S}_b^{s}=\{\bm{s}_i^s\}_{i=1}^n$ through the DA model. The RDS then utilizes the policy $\pi(\bm{S}_b^{s})$ to determine the actions $\bm{A}_b^{s}=\{a_i^s\}_{i=1}^n$ taken on source samples, where $a_i^{s} \in \{0,1\}$. $a_i^{s}=0$ means to filter outlier sample from $\bm{X}_b^s$. Thus, we get a new source batch $\bm{X}_b^{s'}$ related to target domain. Instead of $\bm{X}_b^s$, we feed $\bm{X}_b^{s'}$ into the DA model to solve the PDA problem. Finally, the DA model moves to the next state $s'$ after updated with $\bm{X}_b^{s'}$ and $\bm{X}_b^{t}$, and provides a reward $r_b$ according to the source reconstruction errors based on $G_t$ to update $\pi$ and $V$. In the following sections, we will give a detailed introduction to the state, action, and reward.

\textbf{State:} \label{State} State is defined as a vector $\bm{s}_i^s \in \mathbb{R}^l $. In order to simultaneously consider the unique information of each source sample and the label distribution of target domain when taking action, $\bm{s}_i^s$ concatenates the following features: (1) The high-level semantic feature $\bm{z}_i^s$, which is the output of $F$ given $\bm{x}_i^s$, i.e., $\bm{z}_i^s=F(\bm{x}_i^s)$. (2) The label $y_i^s$ of the source sample, represented by a one-hot vector. (3) The predicted probability distribution $\alpha$ of the target batch $\bm{X}_b^t$, which can be calculated as $\frac{1}{n}\sum_{i=1}^{n}\hat{y}_{i}^t$, $\hat{y}_i^t=C(F(x_i^t))$. Feature (1) represents high-level information of source sample. Feature (3) based on the intuition that the probabilities of assigning the target data to outlier source classes should be small since the target sample is significantly dissimilar to the outlier source sample. Consequently, $\alpha$ quantifies the contribution of each source class to the target domain. Feature (2) is combined with feature (3) to measure the relation between each source sample and the target domain.

\textbf{Action:} The action $a\in \{0,1\}$, which indicates whether the source sample is kept or filtered from the source batch. The selector utilizes  $\epsilon$-greedy strategy \cite{mnih2015human} to sample $a$ based on $\pi(\bm{s}_i^s)$. $\pi(\bm{s}_i^s) \in \mathbb{R}^ {1}$ represents the probability that the sample is kept. The $\epsilon$ is decayed from 1 to 0 as the training progresses. $\pi$ is defined as a policy network with two fully connected layers. Formally, $\pi(\bm{s}_i^s)$ is computed as:
\begin{equation}
\pi(\bm{s}_i^s)=sigmoid(\bm{W}_2\delta(\bm{W}_1\bm{s}_i^s+\bm{b}_1)+\bm{b}_2),
\end{equation}
where $\delta$ is the ReLU activation, $\bm{W}_k$ and $\bm{b}_k$ are the weight matrix and bias of the $k$-th layer, and $\bm{s}_i^s$ is the state of the source sample, which concatenates feature (1), (2) and (3).

\textbf{Reward:}
The selector takes actions to select $\bm{X}_b^{s'}$ from $\bm{X}_b^{s}$. The RTNet uses $\bm{X}_b^{s'}$ to update the DA model and obtains a reward $r_b$ for evaluating the policy. In contrast to usual reinforcement learning, where one reward corresponds to one action, The RTNet assigns one reward to a batch of actions to improve the efficiency of model training.

To take advantage of pixel-level information when selecting source samples, the novel reward is designed according to the reconstruction error of the selected source sample based on $G_t$. The intuition of using this reconstruction error as reward is that the reconstruction error $\|\bm{x}_i^{s'}-G_{t}F(\bm{x}_i^{s'})\|_2^2$ of outlier source sample is large since they are extremely dissimilar to the target classes. Hence, the selector aims to select source samples with small reconstruction errors for distribution alignment and classifier training. However, the purpose of RL is to maximize the reward, so we design the following novel reward based on reconstruction error:
\begin{equation}
\label{e4}
r_{b}=\exp (-{\frac{1}{n'}\sum_{i=1}^{n'}\|\bm{x}_i^{s'}-G_{t}F(\bm{x}_i^{s'})\|_2^2}),
\end{equation}
where $\bm{x}_i^{s'}$ is the sample selected by the reinforced data selector, and $n'$ is the number of samples selected. As shown in Eq. \ref{e4}, the smaller the reconstruction error, the greater the reward, which is in line with our expectations. Note that, to accurately evaluate the efficacy of $\bm{X}_b^{s'}$, rewards are collected after the feature extractor $F$ and classifier $C$ are updated as in Eq. \ref{e5} and before the generators $G_{s,t}$ are updated as in Eq. \ref{e6}. $F$, $C$ and $G_{s,t}$ can be trained as follows:
\begin{small}
\begin{gather}
\label{e5}
\mathop{min}\limits_{(F,C)} \mathcal{L}_{DA}, \\
\label{e6}
\mathop{min}\limits_{(G_{s,t}, F)} \frac{1}{n'}\sum_{i=1}^{n'}\|\bm{x}_i^{s'}-G_{s}F(\bm{x}_i^{s'})\|_2^2+\frac{1}{n}\sum_{i=1}^{n}\|\bm{x}_i^{t}-G_{t}F(\bm{x}_i^{t})\|_2^2.
\end{gather}
\end{small}

In the process of selection, not only the last action contributes to the reward, but all previous actions contribute. Therefore, the future total reward $r'_{b}$ for each batch $b$ can be formalized as:
\begin{equation}
\label{e7}
r'_{b}=\sum_{j=0}^{N-b}\gamma^{j}r_{b+j},
\end{equation}
where $\gamma$ is the reward discount factor, and $N$ is the number of batches in this episode.
\begin{algorithm}[!b]
\caption{The optimization strategy of the RTNet}
\label{alg::algorithm1}
\begin{algorithmic}[1]
\Require
episode number $L$,
source data $\{\bm{X}^s,Y^s\}$
and target data $\bm{X}^t$.
\State Initialize each module in the RTNet.
\For{$episode = 1 \to L$}
\For{\textbf{each} $(\bm{X}_b^{s}, Y_b^s), (\bm{X}_b^{t}) \in (\bm{X}^{s}, Y^s), (\bm{X}^{t})$}
\State Obtain the states $\bm{S}_b^{s}=\{\bm{s}_i^s\}_{i=1}^n$ through the domain adaptation model, where $\bm{s}_i^{s}=[F(x_i^{s}),y_i^{s},\alpha]$.
\State Utilizes  $\epsilon$-greedy strategy to sample $A_b^{s}=\{a_i^s\}_{i=1}^n$ based on $\pi(\bm{S}_b^{s})$.
\State Select source training batch $(\bm{X}_b^{s'}, Y_b^{s'})$ from $(\bm{X}_b^{s}, Y_b^s)$ according to $A_b^{s}$.
\State Update domain adaptation model ($F$ and $C$) with $(\bm{X}_b^{s'}, Y_b^{s'})$ and $(\bm{X}_b^{t})$ as in Eq. \ref{e5}.
\State Obtain reward $r_b$ on $G_t$ with $\bm{X}_b^{s'}$ as in Eq. \ref{e4}.
\State Update $G_{s,t}$ with $\bm{X}_b^{s'}$ and $\bm{X}_b^{t}$ as in Eq. \ref{e6}.
\State Store $(\bm{S}_b^{s}, A_b^{s}, r_b)$ to an episode history $H$.
\EndFor
\For{\textbf{each} $(\bm{S}_b^{s}, A_b^{s}, r_b) \in H$}
\State Obtain the future total reward $r'_b$ as in Eq. \ref{e7}.
\State Obtain the estimated future total reward $V(\bm{S}_b^s)$.
\State Update $\pi$ as Eq. \ref{e9} and update $V$ as Eq. \ref{e11}.
\EndFor
\EndFor
\end{algorithmic}
\end{algorithm}

\textbf{Optimization:} The selector is optimized based on actor-critic algorithm \cite{konda2000actor}.  In each episode, the selector aims to maximize the expected total reward. Formally,  the objective function is defined as:
\begin{equation}
\begin{split}
\label{e8}
\mathcal{J}(\theta)=\mathbb{E}_{\pi_{\theta}}[\sum_{b=1}^{N}r_b],
\end{split}
\end{equation}
where $\theta$ is the parameter of policy network $\pi$. $\theta$ is updated by performing, typically approximate, gradient ascent on $\mathcal{J}(\theta)$. Formally, the update step of $\pi$ is defined as:
\begin{equation}
\label{e9}
\theta=\theta+l*\frac{1}{n}\sum_{i=1}^nv_{i}\nabla_{\theta}\log(\pi_{\theta}(\bm{s}_i^s)),
\end{equation}
where $l$ is the learning rate, $n$ is the batch size, and $v_i$ is an estimate of the advantage function based on future total reward, which guides the update of $\pi$. Note that, $v_{i}\nabla_{\theta}\log(\pi_{\theta}(\bm{s}_i^s))$ is an unbiased estimate of $\nabla_{\theta}\mathcal{J}(\theta)$ \cite{williams1992simple}. The actor-critic framework combines $\pi$ and $V$ for stable training. In this work, we utilize $V_{\Omega}(\bm{s}_i^s)$ to estimate the expected feature total reward. Hence, the $v_i$ can be considered as an estimate of the advantage of action, which can be defined as follows:
\begin{equation}
\label{e10}
v_i=r'_{b}-V_{\Omega}(\bm{s}_i^s).
\end{equation}
The architecture of the value network $V$ is similar to policy network, except that the final output layer is a regression function. $V$ is designed to estimate the expected feature total reward for each state, which can be optimized by:
\begin{equation}
\label{e11}
\Omega=\Omega-l*\frac{1}{n}\sum_{i=1}^n\nabla_{\Omega}\|r'_{b}-V_{\Omega}(\bm{s}_i^s)\|_2^2,
\end{equation}
where $\Omega$ is the trainable parameters of value network $V$.

As the RDS and DA model interact with each other during training, we train them jointly. To ensure that the DA model can provide accurate states and rewards in the early stages of training, we first pre-train $G_{s,t}$, $F$, and $C$ through the classification loss $\mathcal{L}_{s}$ of source samples and  Eq. \ref{e6}. We follow the previous work \cite{qu2019learning} to train the RTNet, the detailed training process is shown in Algorithm \ref{alg::algorithm1}.
\subsection{Theoretical Analysis}
In this section, we prove theoretically that our method improves the expected error boundary on the target sample by using the theory of domain adaptation \cite{ben2010theory}.
\begin{theorem}
Let $\mathcal{H}$ be the common hypothesis class for source $\mathcal{S}$ and target $\mathcal{T}$, the upper bound of the expected error for the target domain, $\epsilon_{t}(h)$,  is defined as:
\begin{equation}
\label{et}
\epsilon_{t}(h)\leq\epsilon_{s}(h)+\frac{1}{2}d_{\mathcal{H}\Delta\mathcal{H}}(p,q)+C, \forall h \in \mathcal{H},
\end{equation}
where the expected error for the target domain is bounded by three terms: (1) $\epsilon_{s}(h)$ is the expected error for source domain; (2) $d_{\mathcal{H}\Delta\mathcal{H}}(p,q)$ is the domain divergence measured by a discrepancy distance between source distribution $p$ and target distribution $q$;  (3) $C=min_{h}[\epsilon_{s}(h)+\epsilon_{t}(h)]$ is the shared error of the ideal joint hypothesis.
\end{theorem}
In Eq. \ref{et}, $\epsilon_{s}(h)$ is expected to be small due to it can be optimized by a deep network with the source labels. Prior DA methods \cite{tzeng2014deep,long2015learning,sun2016deep,ganin2016domain} seek to minimize $d_{\mathcal{H}\Delta\mathcal{H}}(p,q)$  by aligning the global distributions of $\mathcal{S}$ and $\mathcal{T}$. However, Eq. \ref{et} assumes that the label space of the source and target domains is consistent, which is not held in the PDA scenario. Therefore, blindly aligning the global distribution is an erroneous solution, which forces the target sample to align with the outlier source classes (Figure \ref{Fig1}a), resulting in a large $\epsilon_{t}(h)$ in $C$ and triggering a negative transfer. To this end, we need to ensure the consistency of the source and target label spaces. However, it is not possible to directly filter the outlier source classes as the label space of the target domain is unknown. Hence, we propose the RTNet, which extends the DA methods to automatically filter the outlier source classes, so that the Eq. \ref{et} can get the correct results.

\begin{table*}[h]
  \centering
  \caption{Performance on Office-31 dataset and Digital Dataset. RTNet$_{adv}$ represents the model that integrates the reinforced data selector into the DANN. For the integrity and readability of the paper, RTNet$_{adv}$ will be introduced in the Appendix.}
  \label{tab1}\resizebox{\linewidth}{!}{
    \begin{tabular}{cccccccccccccc}
    \toprule
    \multirow{2}{*}{Type}&\multirow{2}{*}{Method}&
    \multicolumn{6}{c}{Office-31}&\multicolumn{6}{c}{Digital Dataset}\cr
    \cmidrule(lr){3-9} \cmidrule(lr){10-14}
    &&A31$\rightarrow$W10&D31$\rightarrow$W10&W31$\rightarrow$D10&A31$\rightarrow$D10&D31$\rightarrow$A10&W31$\rightarrow$A10&Avg&\tabincell{c}{SVHN10 \\$\rightarrow$ MNIST5}&\tabincell{c}{MNIST10 \\ $\rightarrow$MNIST-M5}&\tabincell{c}{USPS10 \\$\rightarrow$MNIST5}&\tabincell{c}{SYN10 \\$\rightarrow$MNIST5}&Avg\cr
    \midrule
    NoA&ResNet / LeNet&76.5$\pm$0.3&$99.2\pm$0.2&$97.7\pm$0.1&$87.5\pm$0.2&$87.2\pm$0.1&$84.1\pm$0.3&88.7&79.6$\pm$0.3&60.2$\pm$0.4&$76.6\pm$0.6&91.3$\pm$0.4&76.9\cr
    \midrule
    \multirow{4}{*}{DA}&DAN\cite{long2015learning} &53.6$\pm$0.7&$62.7\pm$0.5&$57.8\pm$0.6&$47.7\pm$0.5&$61.2\pm$0.6&$69.7\pm$0.5&58.8&63.5$\pm$0.5&$48.9\pm$0.5&$61.3\pm$0.4&55.0$\pm$0.3&57.2\cr
    &DANN\cite{ganin2016domain}&62.8$\pm$0.6&71.6$\pm$0.4&$65.6\pm$0.5&$65.1\pm$0.7&$78.9\pm$0.3&$79.2\pm$0.4&70.5&68.9$\pm$0.7&50.6$\pm$0.7&83.3$\pm$0.5&77.6$\pm$0.4&70.1\cr
    &CORAL\cite{sun2016deep}&52.1$\pm$0.5&65.2$\pm$0.2&$64.1\pm$0.7&$58.0\pm$0.5&$73.1\pm$0.4&$77.9\pm$0.3&65.1&60.8$\pm$0.6&43.4$\pm$0.5&61.7$\pm$0.5&74.4$\pm$0.4&60.1\cr
    &JDDA\cite{chen2019joint}&73.5$\pm$0.6&$93.1\pm$0.3&$89.3\pm$0.2&$76.4\pm$0.4&$77.6\pm$0.1&82.8$\pm$0.2&82.1&72.1$\pm$0.4&54.3$\pm$0.2&71.7$\pm$0.4&85.2$\pm$0.2&70.8\cr
    \midrule
    \multirow{4}{*}{PDA}&PADA \cite{cao2018partial2}&86.3$\pm$0.4&$99.3\pm$0.1&\textbf{100}$\pm$0.0&$90.4\pm$0.1&$91.3\pm$0.2&$92.6\pm$0.1&93.3&90.4$\pm$0.3&89.1$\pm$0.2&$97.4\pm$0.3&$96.5\pm$0.1&93.4\cr
    &ETN\cite{cao2019learning}&93.4$\pm$0.3&$99.3\pm$0.1&$99.2\pm$0.2&$95.5\pm$0.4&\textbf{95.4}$\pm$0.1&$91.7\pm$0.2&95.8&93.6$\pm$0.2&92.5$\pm$0.1&$96.5\pm$0.1&$97.8\pm$0.2&95.1\cr
    \cline{2-14}
    &\textbf{RTNet}&95.1$\pm$0.3&\textbf{100}$\pm$0.0&\textbf{100}$\pm$0.0&$\textbf{97.8}\pm$0.1&$93.9\pm$0.1&$94.1\pm$0.1&96.8&95.3$\pm$0.1&94.2$\pm$0.2&\textbf{98.9}$\pm$0.1&99.2$\pm$0.0&96.9\cr
    &\textbf{RTNet$_{adv}$}&\textbf{96.2}$\pm$0.3&\textbf{100}$\pm$0.0&\textbf{100}$\pm$0.0&$97.6\pm$0.1&$92.3\pm$0.1&$\textbf{95.4}\pm$0.1&\textbf{96.9}&\textbf{97.2}$\pm$0.1&\textbf{94.6}$\pm$0.2&98.5$\pm$0.1&\textbf{99.7}$\pm$0.0&\textbf{97.5}\cr
    \bottomrule
    \end{tabular}}
\end{table*}
\begin{table*}[ht]
\centering
\caption{Performance on the Office-Home dataset. RTNet$_{adv}$ represents the model that integrates the reinforced data selector into the DANN. For the integrity and readability of the paper, RTNet$_{adv}$ will be introduced in the Appendix.}\label{tab:aStrangeTable}
\label{tab2}\resizebox{\linewidth}{!}{
\begin{tabular}{ccccccccccccccc}
\toprule
Type&Method&Ar$\rightarrow$Cl&Ar$\rightarrow$Pr&Ar$\rightarrow$Rw&Cl$\rightarrow$Ar&Cl$\rightarrow$Pr&Cl$\rightarrow$Rw&Pr$\rightarrow$Ar& Pr$\rightarrow$Cl&Pr$\rightarrow$Rw&Rw$\rightarrow$Ar&Rw$\rightarrow$Cl&Rw$\rightarrow$Pr&Avg\\
\midrule
NoA&ResNet &47.2$\pm$0.2&66.8$\pm$0.3&$76.9\pm$0.5&57.6$\pm$0.2&58.4$\pm$0.1&62.5$\pm$0.3&59.4$\pm$0.3&40.6$\pm$0.2&75.9$\pm$0.3&65.6$\pm$0.1&49.1$\pm$0.2&75.8$\pm$0.4&61.3\\
\hline
\multirow{4}{*}{DA}&DAN\cite{long2015learning} &35.7$\pm$0.2&$52.9\pm$0.4&$63.7\pm$0.2&45.0$\pm$0.3&51.7$\pm$0.3&49.3$\pm$0.1&42.4$\pm$0.2&31.5$\pm$0.4&68.7$\pm$0.1&59.7$\pm$0.3&34.6$\pm$0.4&67.8$\pm$0.1&50.3\\
&DANN\cite{ganin2016domain} &43.2$\pm$0.5&61.9$\pm$0.2&72.1$\pm$0.4&52.3$\pm$0.4&53.5$\pm$0.2&57.9$\pm$0.1&47.2$\pm$0.3&35.4$\pm$0.1&70.1$\pm$0.3&61.3$\pm$0.2&37.0$\pm$0.2&71.7$\pm$0.3&55.3\\
&CORAL\cite{sun2016deep} &38.2$\pm$0.1&55.6$\pm$0.3&65.9$\pm$0.2&48.4$\pm$0.4&52.5$\pm$0.1&51.3$\pm$0.2&48.9$\pm$0.3&32.6$\pm$0.1&67.1$\pm$0.2&63.8$\pm$0.4&35.9$\pm$0.2&69.8$\pm$0.1&52.5\\
&JDDA\cite{chen2019joint} &45.8$\pm$0.4&63.9$\pm$0.2&74.1$\pm$0.3&51.8$\pm$0.2&55.2$\pm$0.3&60.3$\pm$0.2&53.7$\pm$0.2&38.3$\pm$0.1&72.6$\pm$0.2&62.5$\pm$0.1&43.3$\pm$0.3&71.3$\pm$0.1&57.7\\
\hline
\multirow{3}{*}{PDA}&PADA\cite{cao2018partial2} &53.2$\pm$0.2&69.5$\pm$0.1&$78.6\pm$0.1&$61.7\pm$0.2&62.7$\pm$0.3&60.9$\pm$0.1&56.4$\pm$0.5&44.6$\pm$0.2&79.3$\pm$0.1&74.2$\pm$0.1&55.1$\pm$0.3&77.4$\pm$0.2&64.5\\
&ETN\cite{cao2019learning} &60.4$\pm$0.3&76.5$\pm$0.2&$77.2\pm$0.3&$64.3\pm$0.1&67.5$\pm$0.3&75.8$\pm$0.2&69.3$\pm$0.1&54.2$\pm$0.1&83.7$\pm$0.2&75.6$\pm$0.3&56.7$\pm$0.2&84.5$\pm$0.3&70.5\\
\cline{2-15}
&\textbf{RTNet} &62.7$\pm$0.1&79.3$\pm$0.2&\textbf{81.2}$\pm$0.1&65.1$\pm$0.1&68.4$\pm$0.3&76.5$\pm$0.1&70.8$\pm$0.2&\textbf{55.3}$\pm$0.1&\textbf{85.2}$\pm$0.3&76.9$\pm$0.2&\textbf{59.1}$\pm$0.2&83.4$\pm$0.3&72.0\\
&\textbf{RTNet$_{adv}$} &\textbf{63.2}$\pm$0.1&\textbf{80.1}$\pm$0.2&80.7$\pm$0.1&\textbf{66.7}$\pm$0.1&\textbf{69.3}$\pm$0.2&\textbf{77.2}$\pm$0.2&\textbf{71.6}$\pm$0.3&53.9$\pm$0.3&84.6$\pm$0.1&\textbf{77.4}$\pm$0.2&57.9$\pm$0.3&\textbf{85.5}$\pm$0.1&\textbf{72.3}\\
\bottomrule
\end{tabular}}
\end{table*}
\section{Experiments}\label{section4}
\subsection{Datasets}
\textbf{Office-31}\cite{saenko2010adapting} is a widely-used visual domain adaptation dataset, which contains 4,110 images of 31 categories from three distinct domains:  Amazon website (A), Webcam (W) and  DSLR camera (D).  Following the settings in \cite{cao2018partial}, we select the same 10 categories in each domain to build new target domains and create 6 transfer scenarios as in Table \ref{tab1}.

\textbf{Digital Dataset} includes five domain adaptation benchmarks: Street View House Numbers (SVHN) \cite{Netzer2011Reading}, MNIST \cite{Lecun1998Gradient}, MNIST-M \cite{ganin2016domain}, USPS \cite{Hull2002A} and synthetic digits dataset (SYN) \cite{ganin2016domain}, which consist of ten categories. We select 5 categories (digit 0 to digit 4) as target domain in each dataset and construct four PDA tasks as in Table \ref{tab1}.

\textbf{Office-Home}\cite{venkateswara2017deep} is a more challenging DA dataset, which consists of 4 different domains: Artistic images (Ar), Clipart images (Cl), Product images (Pr) and Real-World images (Rw). For each transfer task, when a domain is used as the source domain, we use samples from all 65 categories; when a domain is used as the target domain, we select the samples from the same 25 categories as \cite{cao2019learning}. Hence, we can build twelve PDA tasks as in Table \ref{tab2}.
\begin{figure*}[ht]
    \centering
  \begin{subfigure}[b]{0.24\textwidth}
    \includegraphics[width=\textwidth]{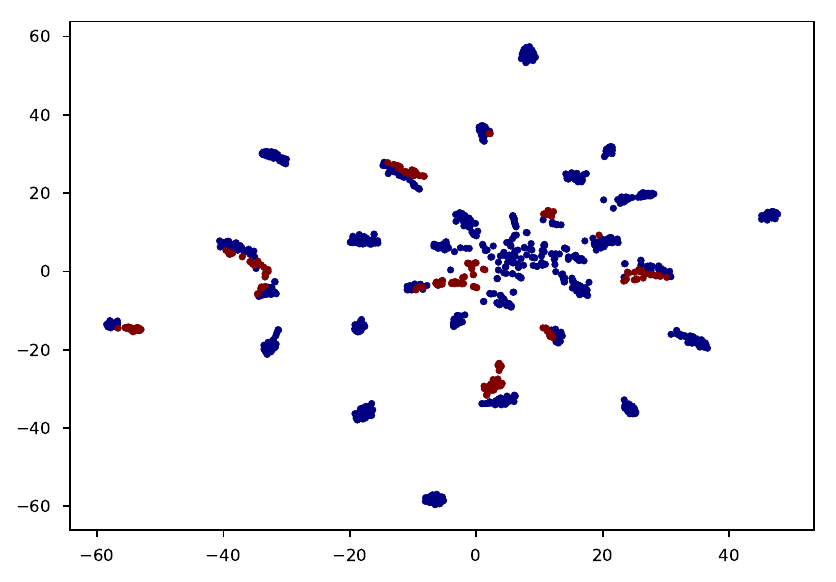}
    \caption{ResNet-50}
    \label{2D1}
  \end{subfigure}
   \begin{subfigure}[b]{0.24\textwidth}
    \includegraphics[width=\textwidth]{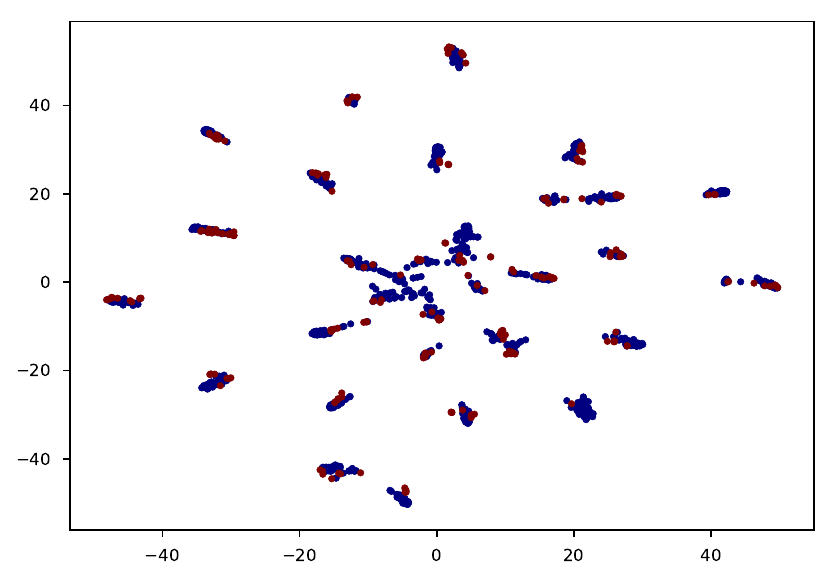}
    \caption{CORAL}
    \label{2D2}
  \end{subfigure}
    \begin{subfigure}[b]{0.24\textwidth}
    \includegraphics[width=\textwidth]{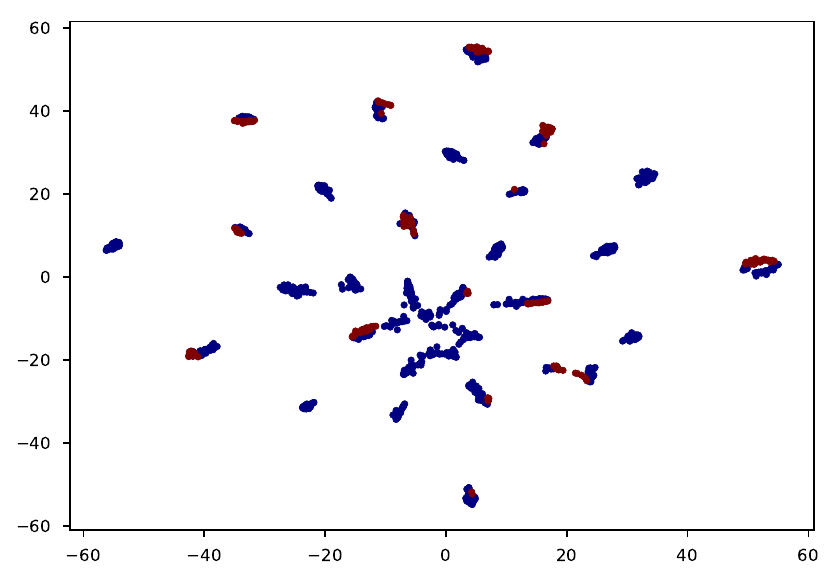}
    \caption{ETN}
    \label{2D3}
  \end{subfigure}
    \begin{subfigure}[b]{0.24\textwidth}
    \includegraphics[width=\textwidth]{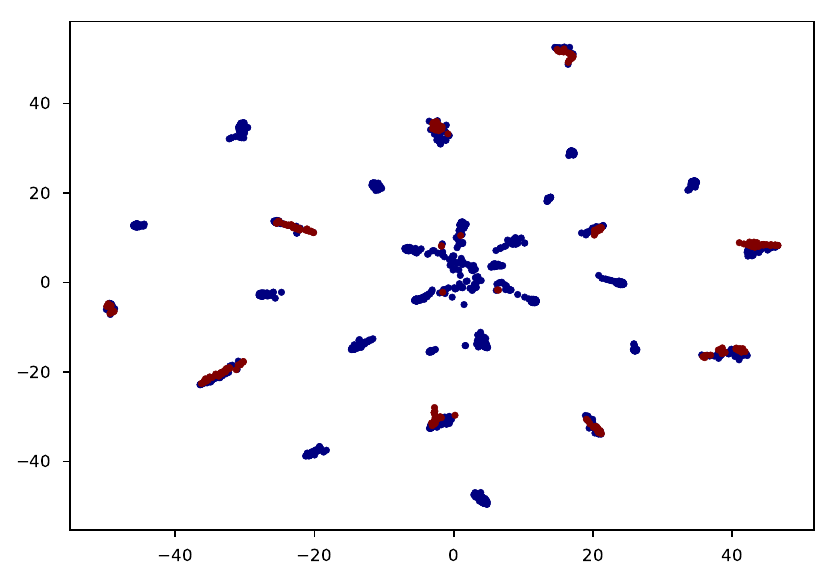}
    \caption{RTNet}
    \label{2D4}
  \end{subfigure}
     \begin{subfigure}[b]{0.24\textwidth}
    \includegraphics[width=\textwidth]{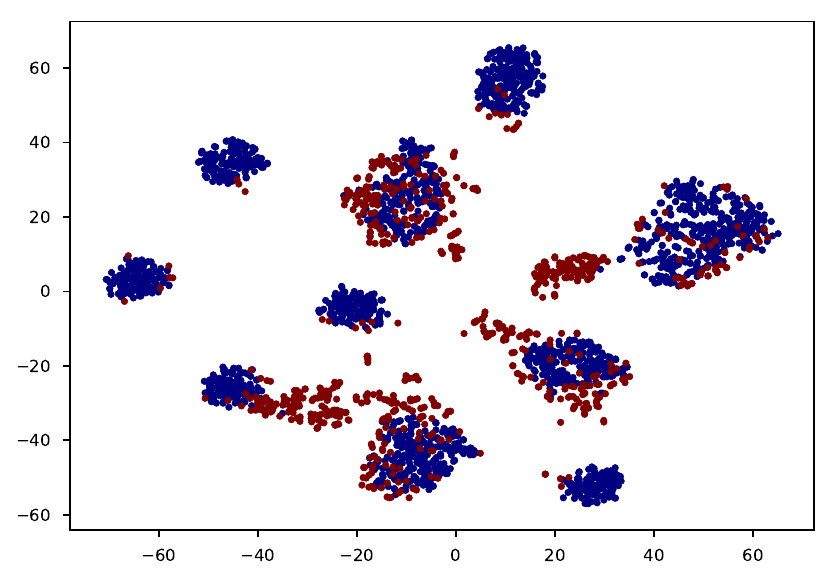}
    \caption{LeNet}
    \label{fig:1}
  \end{subfigure}
  \begin{subfigure}[b]{0.24\textwidth}
    \includegraphics[width=\textwidth]{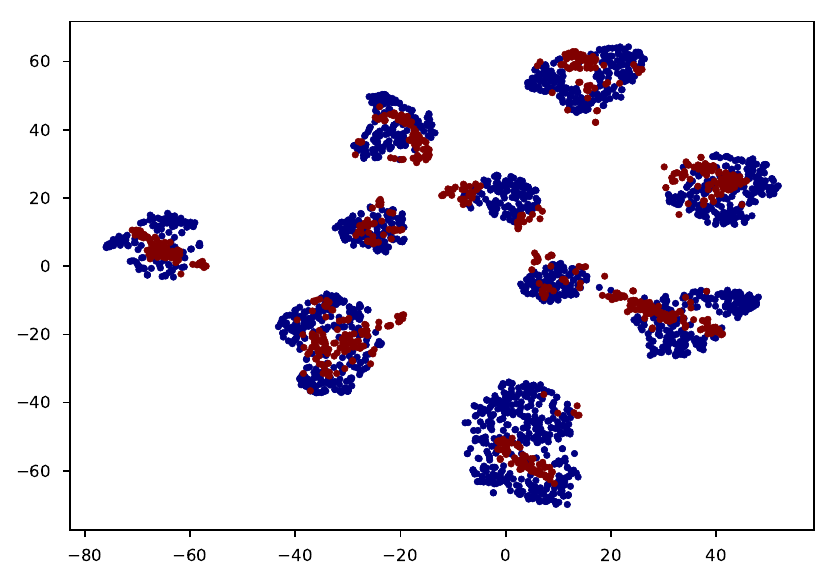}
    \caption{CORAL}
    \label{fig:2}
  \end{subfigure}
    \begin{subfigure}[b]{0.24\textwidth}
    \includegraphics[width=\textwidth]{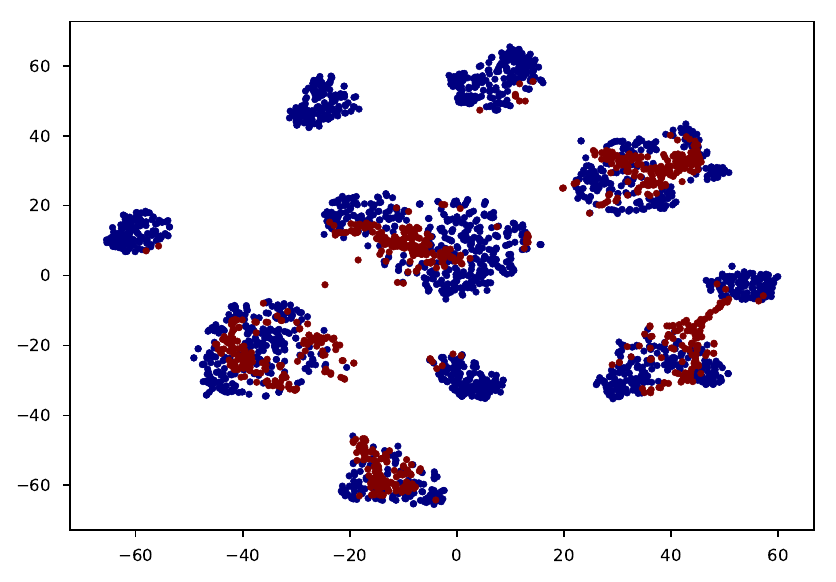}
    \caption{ETN}
    \label{fig:3}
  \end{subfigure}
  \begin{subfigure}[b]{0.24\textwidth}
    \includegraphics[width=\textwidth]{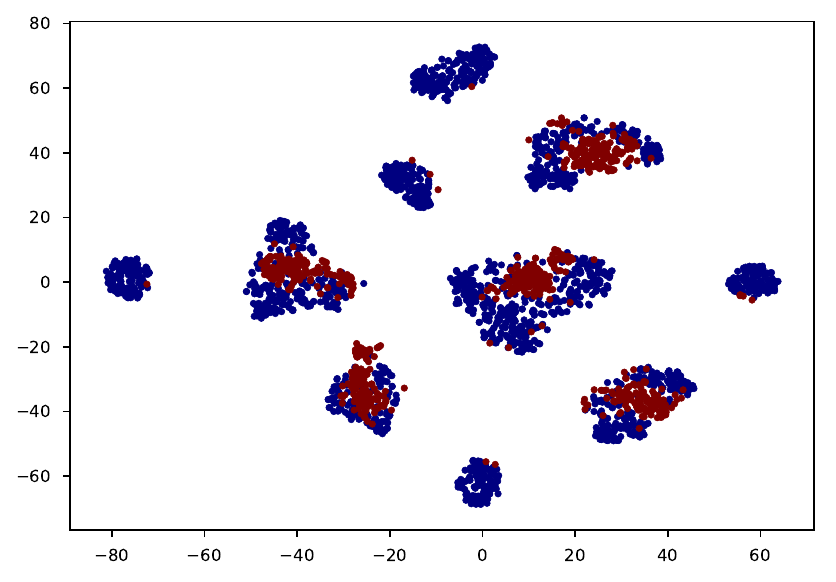}
    \caption{RTNet}
    \label{fig:4}
  \end{subfigure}
\caption{ The t-SNE visualization on A31$\rightarrow$W10 ((a)-(d)) and SVHN10$\rightarrow$MNIST5 ((e)-(h)). Red points represent target samples and blue points represent source samples. The results generated from category information are shown in Appendix.}
\label{tsne}
\end{figure*}
\subsection{Implementation Details}
The RTNet is implemented via Tensorflow and trained with the Adam optimizer. For the experiments on Office-31 and Office-Home, we employ the ResNet-50 pre-trained on ImageNet as the backbone of domain adaptation model and fine-tune the parameters of the fully connected layers and the final block. For the experiments on digital datasets,  we adopt modified LeNet as the backbone of domain adaptation model and update all of the weights. All images are converted to grayscale and resized to 32 $\times$ 32.

In RTNet, the structure of each module can be seen in Appendix. To guarantee fair comparison, the same frameworks are used for $F$ and $C$ in all comparison methods, and each method is trained five times and the average is taken as the final result. For all hyperparameters, we set $l=1e-4$, $\lambda_1=7$ and $\gamma=0.85$, which selected by using a grid search on the performance of the validation set. The parameter sensitivity analysis can be seen in Appendix. To ease model selection, the hyperparameters of comparison methods are gradually changing from 0 to 1 as in  \cite{long2017deep}.
\subsection{Result and Discussion}\label{result}
Tables \ref{tab1} and \ref{tab2} show the classification results on three datasets. By looking these tables, several observations can be made. (1) the previous standard DA methods including those based on adversarial network (DANN), and those based on moment match (DAN, JDDA, and CORAL) perform even worse than non-adaptation (NoA) model, indicating that they were affected by the negative transfer. (2) PDA methods (ETN and PADA) improve classification accuracy by a large margin since their weighting mechanisms can mitigate negative transfer caused by outlier categories. (3) Comparing the model with the RDS (RTNet and RTNet$_{adv}$) and the model without the RDS (CORAL and DANN), the model with the RDS can alleviate the negative transfer to greatly improve the performance of the model in the target domain. This proves that the selector we design is a general model and can be easily integrated into existing DA models, including not only matching-based methods but also adversarial-based methods. (4) RTNet / RTNet$_{adv}$ achieves the best accuracy on most transfer tasks. Different from the previous PDA methods which only rely on the high-level information to obtain the weight, RTNet / RTNet$_{adv}$ adopts the high-level information to select the source sample, and employs the pixel-level information as evaluation criteria to guide the learning of policy network. Thus, this selection mechanism can detect outlier source classes more effectively and transfer relevant samples.
\subsection{Analysis}
\textbf{Feature Visualization:}
We visualize the features of adaptation layer using t-SNE \cite{donahue2014decaf}. As shown in Figure \ref{tsne}, several observations can be made. (1) by comparing Figures \ref{2D1}, \ref{fig:1} and Figures \ref{2D2}, \ref{fig:2}, we find that CORAL forces the target domain to be aligned with the whole source domain, including outlier classes that do not exist in the target label space, which triggers negative transfer. (2) as can be seen in Figures \ref{2D4}, \ref{fig:4}, RTNet correctly matches the target samples to related source samples by integrating the RDS into CORAL to filter outlier classes, which confirms that the matching-based DA methods can be extended to solve PDA problem by embedding RDS. (3) compared with Figure \ref{2D3}, \ref{fig:3}, RTNet matches the related source domain and the target domain more accurately, indicating that it is more effective than ETN in suppressing the side effect of outlier classes by considering high-level and pixel-level information.
\begin{figure*}[ht]
    \centering
     \begin{subfigure}[b]{0.185\textwidth}
    \includegraphics[width=\textwidth]{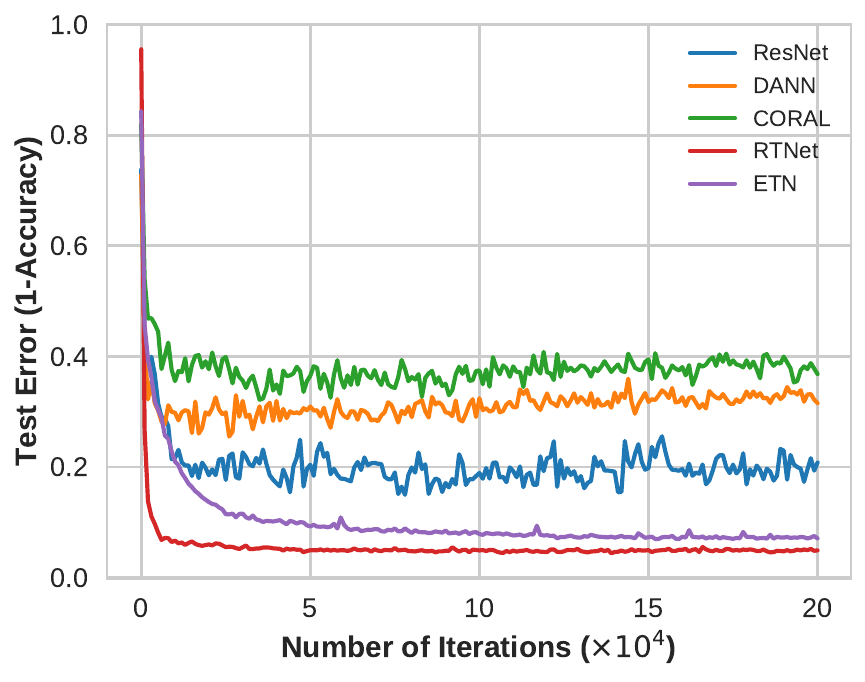}
    \caption{Convergence}
    \label{conv}
  \end{subfigure}
     \begin{subfigure}[b]{0.19\textwidth}
    \includegraphics[width=\textwidth]{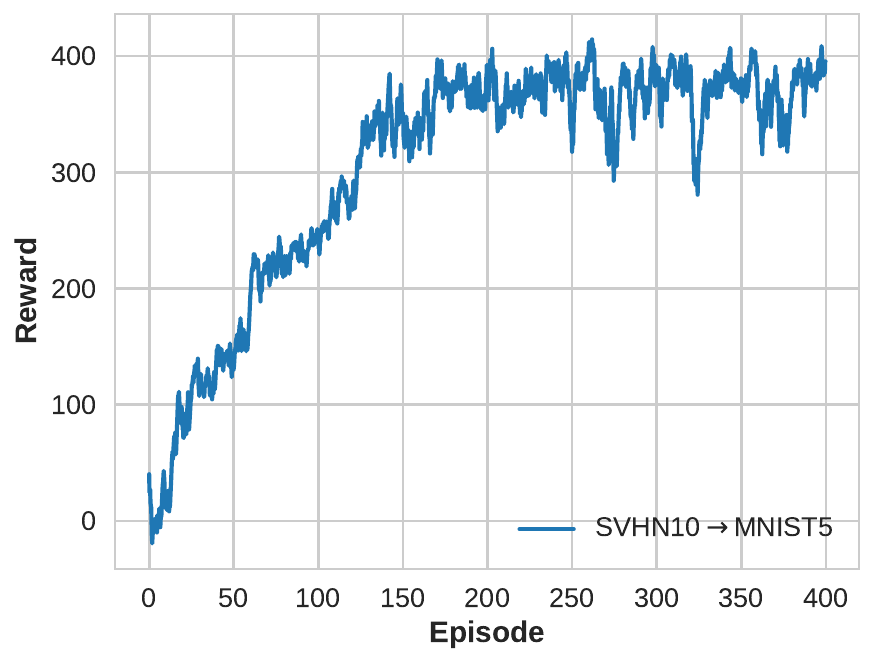}
    \caption{Learning curve}
    \label{lear}
  \end{subfigure}
   \begin{subfigure}[b]{0.19\textwidth}
    \includegraphics[width=\textwidth]{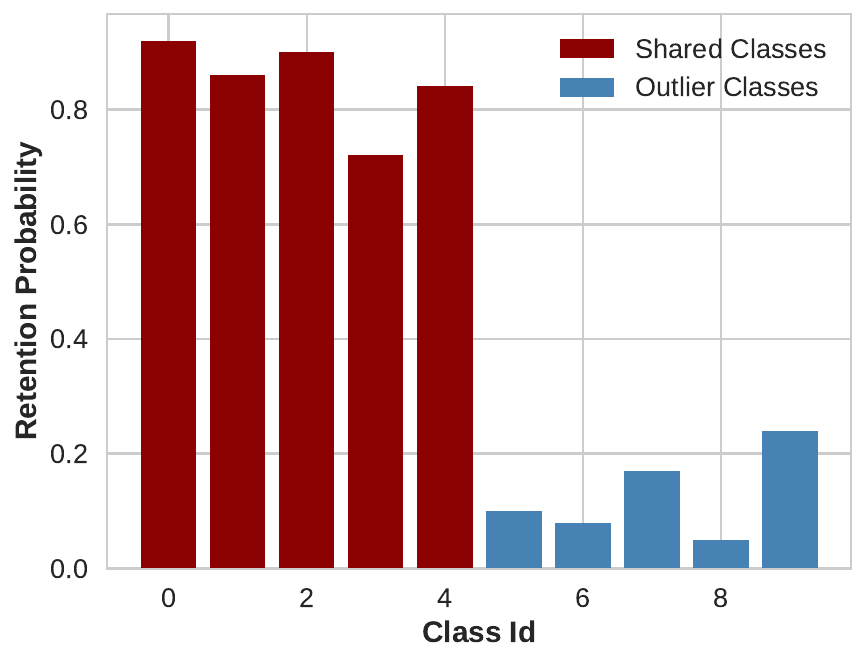}
    \caption{Probability in $\mathcal{C}_s$}
    \label{pro}
  \end{subfigure}
     \begin{subfigure}[b]{0.19\textwidth}
    \includegraphics[width=\textwidth]{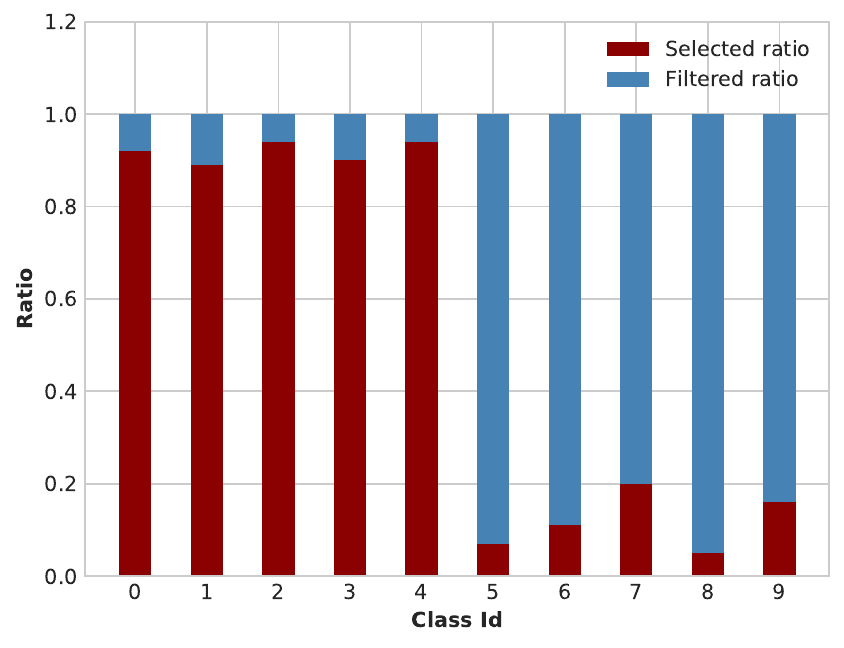}
    \caption{Select \& filter ratio}
    \label{ratio}
  \end{subfigure}
   \begin{subfigure}[b]{0.19\textwidth}
    \includegraphics[width=\textwidth]{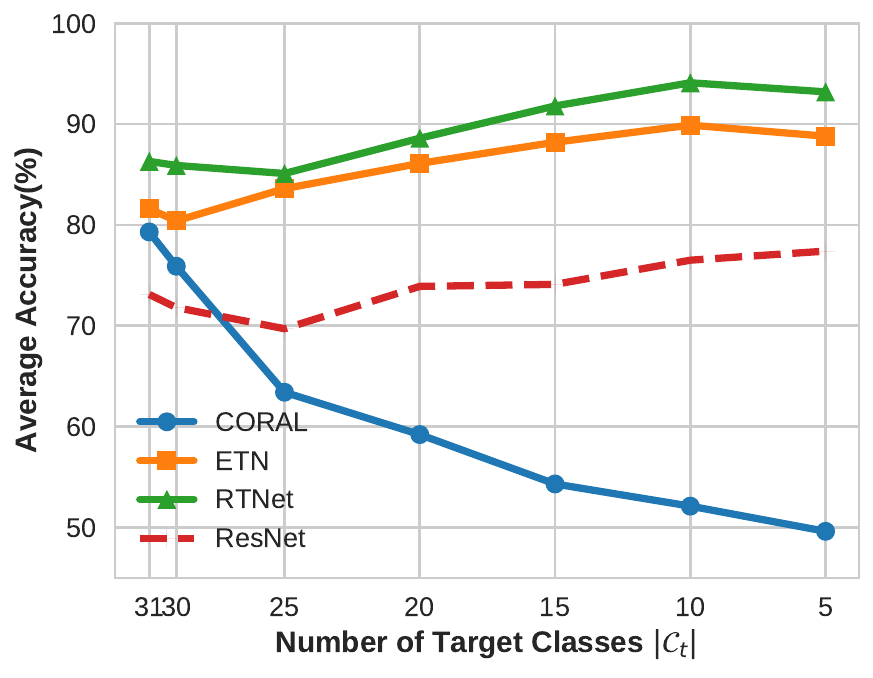}
    \caption{Acc w.r.t. $|\mathcal{C}_{t}|$ }
    \label{sen}
  \end{subfigure}
\caption{(a) Convergence analysis on SVHN10$\rightarrow$MNIST5. (b) Learning curve on SVHN10$\rightarrow$MNIST5. (c) Source class-wise retention probability learned by policy network on SVHN10$\rightarrow$MNIST5. (d) Class-wise selected ratio and filtered ratio evaluated by RDS trained on SVHN10$\rightarrow$MNIST5. (e) The accuracy curve of varying the number of target classes on A$\rightarrow$W.}
\label{analy}
\end{figure*}

\textbf{Convergence Performance:}
We analyze the convergence of RTNet. As shown in Figure \ref{conv}, the test errors of DANN and CORAL are higher than ResNet due to negative transfer. RTNet fast and stably converges to the lowest test error, indicating that it can be efficiently trained to solve PDA problem. As shown in Figure \ref{lear}, the reward gradually increases as the episode progresses, meaning that the RDS can learn the correct policy to maximize the reward and filter out outlier source classes.
\subsection{Case Study and Performance Interpretation}
The results in Section \ref{result} demonstrate the effectiveness of the RTNet. However, the lack of interpretability of the neural architecture makes it difficult to speculate on the reasons behind decisions made by RDS. Therefore, we introduce the overall performance and specific case to prove the ability of the selector to select and filter samples.

\textbf{Statistics of Class-wise Retention Probabilities:}
We utilize $ E(\pi_{\theta}(\bm{S}_{c}^s))$ to verify the ability of the selector to filter the samples,   averaging the retention probabilities of each class of source domain. $\bm{S}_{c}^s$ represents a source sample set, which contains samples belonging to class $c$. As shown in Figure \ref{pro}, RTNet assigns much larger retention probabilities to the shared classes than to the outlier classes. These results prove that RTNet has the ability to automatically select relevant source classes and filter out outlier classes. Besides, the outlier classes with a similar appearance to the shared classes, such as 7 and 9,  have a larger retention probability than other outlier classes, which indicates that the selector we develop can indeed select source samples similar to the target domain based on the pixel-level information.

\textbf{Class-wise Selected Ratio and Filtered Ratio:}
 We input the sampled SVHN samples into RTNet for sample selection.  As shown in Figure \ref{ratio}, the outlier samples (5, 6 and 8) that differ significantly in appearance from the shared samples (0-4) can be filtered out by 92\% on average, while the outlier classes (7 and 9) with smaller appearance differences from the shared classes can be filtered out by 72\% on average. For shared classes, the ratio of samples filtered in each class is not much different, and an average of 8.2\% of the samples are filtered by error.  These results indicate that RTNet can effectively filter outlier source samples, especially those that have large differences in appearance with shared classes and keep related source samples well.

\textbf{Wasserstein Distances between Domains:} The Wasserstein distance measures the distance between two probability distributions \cite{Villani2003Topics}. We take the selection of the selector in the last episode as the result of the selection and calculate the Wasserstein distance between target samples and source samples, including selected and filtered source samples. We observe that the patterns of the two tasks are identical through the results of Table \ref{wass}:  (1) $W_{select}<W_{all}$, which indicates that the sample selected by the selector is closer to the target domain and thus may contribute to the transfer process. (2) $W_{filter}>W_{all}$, which means that the filtered source samples are extremely dissimilar to the target domain and may result in a negative transfer. These findings show that our proposal can select source samples whose Wasserstein distances are close to the target domain. This makes sense because such source samples can be more easily transferred and helpful to the target domain.
\begin{table}[ht]
\centering
\caption{The Wasserstein distances between domains.}\label{tab:aStrangeTable}
\label{wass}
\begin{tabular}{cccc}
\toprule
Name &Domains&\tabincell{c}{SVHN10\\$\rightarrow$MNIST5}&A31$\rightarrow$W10\\
\midrule
$W_{all}$ &$\mathrm{T}\leftrightarrow\mathrm{S}$&0.2574&4.3233 \\
$W_{select}$&$\mathrm{T}\leftrightarrow\mathrm{S_{select}}$&0.1645&2.3179\\
$W_{filter}$&$\mathrm{T}\leftrightarrow\mathrm{S_{filter}}$&0.3679&5.6308\\
\bottomrule
\end{tabular}
\end{table}

\begin{figure}[!ht]
   \centering
     \includegraphics*[width=0.8\linewidth]{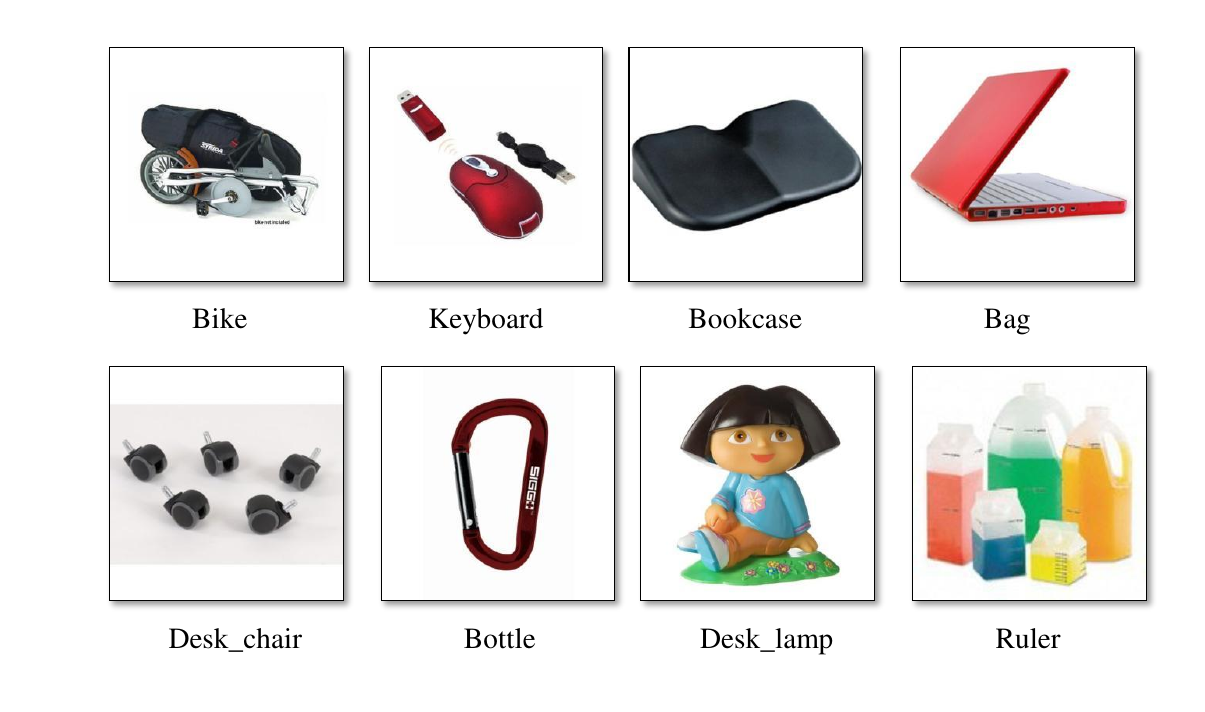}
   \caption{Case study on A31$\rightarrow$W31 task. These noisy samples with mismatches between labels and images are sampled from the source samples filtered by RDS.  The description below the image is the label provided by the dataset.}
   \label{case}
\end{figure}

\textbf{Target Classes:} We conduct experiments to evaluate the performance of RTNet when the number of target classes varies. As shown in Figure \ref{sen}, as the number of target classes decreases, the performance of CORAL degrades rapidly, indicating that negative transfer becomes more and more serious as the label distribution becomes larger.  RTNet performs better than other methods, indicating that it can suppress negative transfer more effectively. Moreover, RTNet is superior to CORAL when the source and target label spaces are consistent (A31$\rightarrow$W31), which shows that our method does not filter erroneously when there are no outlier classes.  As shown in Figure \ref{case},  on the A31$\rightarrow$W31 task, we analyze some of the source samples filtered by RDS and find that most of them are noise samples, which have mismatches between labels and images. For example, an image of a mouse is wrongly labeled as a keyboard in the Office-31 dataset.  This case study shows that the RDS can filter noisy samples to improve the performance even if the label space of the source and target domains are consistent.
%
%
%
\section{Conclusion}
In this work, we propose an end-to-end RTNet, which utilizes both high-level and pixel-level information to address PDA problem. RTNet applies RL to train a reinforced data selector based on actor-critic framework to filter outlier source classes with the purpose of mitigating negative transfer. Unlike previous adversarial-based PDA methods, The RDS we proposed can be integrated into almost all DA models including those based on adversarial network, and those based on moment match. Note that, the results of RTNet$_{adv}$ based on adversarial model are shown in Appendix. The state-of-the-art experimental results confirm the efficacy of RTNet.
{\small
\bibliographystyle{ieee_fullname}
\bibliography{egbib}

\begin{thebibliography}{10}\itemsep=-1pt

\bibitem{arulkumaran2017deep}
Kai Arulkumaran, Marc~Peter Deisenroth, Miles Brundage, and Anil~Anthony
  Bharath.
\newblock Deep reinforcement learning: A brief survey.
\newblock {\em IEEE Signal Processing Magazine}, 34(6):26--38, 2017.

\bibitem{ben2010theory}
Shai Ben-David, John Blitzer, Koby Crammer, Alex Kulesza, Fernando Pereira, and
  Jennifer~Wortman Vaughan.
\newblock A theory of learning from different domains.
\newblock {\em Machine learning}, 79(1-2):151--175, 2010.

\bibitem{candela2009dataset}
J~Qui{\~n}onero Candela, Masashi Sugiyama, Anton Schwaighofer, and Neil~D
  Lawrence.
\newblock Dataset shift in machine learning, 2009.

\bibitem{cao2018partial}
Zhangjie Cao, Mingsheng Long, Jianmin Wang, and Michael~I Jordan.
\newblock Partial transfer learning with selective adversarial networks.
\newblock In {\em CVPR}, pages 2724--2732, 2018.

\bibitem{cao2018partial2}
Zhangjie Cao, Lijia Ma, Mingsheng Long, and Jianmin Wang.
\newblock Partial adversarial domain adaptation.
\newblock In {\em ECCV}, pages 135--150, 2018.

\bibitem{cao2019learning}
Zhangjie Cao, Kaichao You, Mingsheng Long, Jianmin Wang, and Qiang Yang.
\newblock Learning to transfer examples for partial domain adaptation.
\newblock {\em arXiv preprint arXiv:1903.12230}, 2019.

\bibitem{chen2019joint}
Chao Chen, Zhihong Chen, Boyuan Jiang, and Xinyu Jin.
\newblock Joint domain alignment and discriminative feature learning for
  unsupervised deep domain adaptation.
\newblock In {\em AAAI}, volume~33, pages 3296--3303, 2019.

\bibitem{chen2019homm}
Chao Chen, Zhihang Fu, Zhihong Chen, Sheng Jin, Zhaowei Cheng, Xinyu Jin, and
  Xian-Sheng Hua.
\newblock Homm: Higher-order moment matching for unsupervised domain
  adaptation.
\newblock {\em arXiv preprint arXiv:1912.11976}, 2019.

\bibitem{chen2019deep}
Zhihong Chen, Chao Chen, Xinyu Jin, Yifu Liu, and Zhaowei Cheng.
\newblock Deep joint two-stream wasserstein auto-encoder and selective
  attention alignment for unsupervised domain adaptation.
\newblock {\em Neural Computing and Applications}, pages 1--14, 2019.

\bibitem{donahue2014decaf}
Jeff Donahue, Yangqing Jia, Oriol Vinyals, Judy Hoffman, Ning Zhang, Eric
  Tzeng, and Trevor Darrell.
\newblock Decaf: A deep convolutional activation feature for generic visual
  recognition.
\newblock In {\em ICML}, pages 647--655, 2014.

\bibitem{fang2017learning}
Meng {Fang}, Yuan {Li}, and Trevor {Cohn}.
\newblock Learning how to active learn: A deep reinforcement learning approach.
\newblock In {\em EMNLP}, pages 595--605, 2017.

\bibitem{ganin2016domain}
Yaroslav Ganin, Evgeniya Ustinova, Hana Ajakan, Pascal Germain, Hugo
  Larochelle, Fran{\c{c}}ois Laviolette, Mario Marchand, and Victor Lempitsky.
\newblock Domain-adversarial training of neural networks.
\newblock {\em The Journal of Machine Learning Research}, 17(1):2096--2030,
  2016.

\bibitem{hoffman2018cycada}
Judy Hoffman, Eric Tzeng, Taesung Park, Jun-Yan Zhu, Phillip Isola, Kate
  Saenko, Alexei Efros, and Trevor Darrell.
\newblock Cycada: Cycle-consistent adversarial domain adaptation.
\newblock In {\em ICML}, 2018.

\bibitem{Hull2002A}
J.~J. Hull.
\newblock A database for handwritten text recognition research.
\newblock {\em IEEE Transactions on Pattern Analysis \& Machine Intelligence},
  16(5):550--554, 2002.

\bibitem{konda2000actor}
Vijay~R Konda and John~N Tsitsiklis.
\newblock Actor-critic algorithms.
\newblock In {\em NeurIPS}, pages 1008--1014, 2000.

\bibitem{Lecun1998Gradient}
Y.~L. Lecun, Leon Bottou, Yoshua Bengio, and Patrick Haffner.
\newblock Gradient-based learning applied to document recognition. proc ieee.
\newblock {\em Proceedings of the IEEE}, 86(11):2278--2324, 1998.

\bibitem{long2015learning}
Mingsheng Long, Yue Cao, Jianmin Wang, and Michael Jordan.
\newblock Learning transferable features with deep adaptation networks.
\newblock In {\em ICML}, pages 97--105, 2015.

\bibitem{long2017deep}
Mingsheng Long, Han Zhu, Jianmin Wang, and Michael~I Jordan.
\newblock Deep transfer learning with joint adaptation networks.
\newblock In {\em ICML}, pages 2208--2217, 2017.

\bibitem{mnih2015human}
Volodymyr Mnih, Koray Kavukcuoglu, David Silver, Andrei~A Rusu, Joel Veness,
  Marc~G Bellemare, Alex Graves, Martin Riedmiller, Andreas~K Fidjeland, Georg
  Ostrovski, et~al.
\newblock Human-level control through deep reinforcement learning.
\newblock {\em Nature}, 518(7540):529, 2015.

\bibitem{Netzer2011Reading}
Yuval Netzer, Tao Wang, Adam Coates, Alessandro Bissacco, Bo Wu, and Andrew~Y.
  Ng.
\newblock Reading digits in natural images with unsupervised feature learning.
\newblock {\em Nips Workshop on Deep Learning \& Unsupervised Feature
  Learning}, 2011.

\bibitem{pan2010survey}
Sinno~Jialin Pan and Qiang Yang.
\newblock A survey on transfer learning.
\newblock {\em IEEE Transactions on knowledge and data engineering},
  22(10):1345--1359, 2010.

\bibitem{qu2019learning}
Chen Qu, Feng Ji, Minghui Qiu, Liu Yang, Zhiyu Min, Haiqing Chen, Jun Huang,
  and W~Bruce Croft.
\newblock Learning to selectively transfer: Reinforced transfer learning for
  deep text matching.
\newblock In {\em WSDM}, pages 699--707, 2019.

\bibitem{rummery1994line}
Gavin~A Rummery and Mahesan Niranjan.
\newblock {\em On-line Q-learning using connectionist systems}, volume~37.
\newblock University of Cambridge, Department of Engineering Cambridge,
  England, 1994.

\bibitem{saenko2010adapting}
Kate Saenko, Brian Kulis, Mario Fritz, and Trevor Darrell.
\newblock Adapting visual category models to new domains.
\newblock In {\em ECCV}, pages 213--226, 2010.

\bibitem{sun2016deep}
Baochen Sun and Kate Saenko.
\newblock Deep coral: Correlation alignment for deep domain adaptation.
\newblock In {\em ECCV}, pages 443--450, 2016.

\bibitem{tzeng2017adversarial}
Eric Tzeng, Judy Hoffman, Kate Saenko, and Trevor Darrell.
\newblock Adversarial discriminative domain adaptation.
\newblock In {\em CVPR}, volume~1, page~4, 2017.

\bibitem{tzeng2014deep}
Eric Tzeng, Judy Hoffman, Ning Zhang, Kate Saenko, and Trevor Darrell.
\newblock Deep domain confusion: Maximizing for domain invariance.
\newblock {\em arXiv preprint arXiv:1412.3474}, 2014.

\bibitem{venkateswara2017deep}
Hemanth Venkateswara, Jose Eusebio, Shayok Chakraborty, and Sethuraman
  Panchanathan.
\newblock Deep hashing network for unsupervised domain adaptation.
\newblock In {\em CVPR}, pages 5018--5027, 2017.

\bibitem{Villani2003Topics}
Cédric Villani.
\newblock Topics in optimal transportation.
\newblock {\em Ams Graduate Studies in Mathematics}, page 370, 2003.

\bibitem{williams1992simple}
Ronald~J Williams.
\newblock Simple statistical gradient-following algorithms for connectionist
  reinforcement learning.
\newblock {\em Machine learning}, 8(3-4):229--256, 1992.

\bibitem{wu2018reinforced}
Jiawei {Wu}, Lei {Li}, and William~Yang {Wang}.
\newblock Reinforced co-training.
\newblock In {\em NAACL HLT 2018}, pages 1252--1262, 2018.

\bibitem{zhang2018importance}
Jing Zhang, Zewei Ding, Wanqing Li, and Philip Ogunbona.
\newblock Importance weighted adversarial nets for partial domain adaptation.
\newblock In {\em CVPR}, pages 8156--8164, 2018.

\end{thebibliography}
}
\end{document}